\documentclass{article}

\usepackage[final]{neurips}
\usepackage[utf8]{inputenc} 
\usepackage[T1]{fontenc}    
\usepackage{hyperref}       
\usepackage{url}            
\usepackage{booktabs}       
\usepackage{amsfonts}       
\usepackage{nicefrac}       
\usepackage{microtype}      
\usepackage{lipsum}
\usepackage{graphicx}
\usepackage{enumitem}
\usepackage{color}
\usepackage{verbatimbox}
\usepackage{listings}
\usepackage{xcolor}
\usepackage[numbers]{natbib}

\definecolor{codegreen}{rgb}{0,0.6,0}
\definecolor{codegray}{rgb}{0.5,0.5,0.5}
\definecolor{codepurple}{rgb}{0.58,0,0.82}
\definecolor{backcolour}{rgb}{0.95,0.95,0.92}

\setlist[itemize]{leftmargin=*}
\setlist[enumerate]{leftmargin=*}
\graphicspath{ {./images/} }

\lstdefinestyle{mystyle}{
  backgroundcolor=\color{backcolour}, commentstyle=\color{codegreen},
  keywordstyle=\color{magenta},
  numberstyle=\tiny\color{codegray},
  stringstyle=\color{codepurple},
  basicstyle=\ttfamily\footnotesize,
  breakatwhitespace=false,         
  breaklines=true,                 
  captionpos=b,                    
  keepspaces=true,                 
  numbers=left,                    
  numbersep=5pt,                  
  showspaces=false,                
  showstringspaces=false,
  showtabs=false,                  
  tabsize=2
}

\usepackage[strict]{changepage}
\usepackage{xcolor}
\usepackage{framed}
\definecolor{demonstrationshade}{rgb}{0.95,0.95,1}
\definecolor{promptshade}{rgb}{0.95,0.95,1}

\title{NPHardEval: Dynamic Benchmark on Reasoning Ability of Large Language Models via Complexity Classes}

\author{
    Lizhou Fan$^{\dagger*}$, Wenyue Hua$^{\ddagger*}$, Lingyao Li$^\dagger$, Haoyang Ling$^\dagger$, Yongfeng Zhang$^\ddagger$\\
    $^\dagger$School of Information, University of Michigan, Ann Arbor, MI 48103 \\
    $^\ddagger$Department of Computer Science, Rutgers University, New Brunswick, NJ 08854 \\
    lizhouf@umich.edu, wenyue.hua@rutgers.edu, \{lingyaol, hyfrankl\}@umich.edu, \\ yongfeng.zhang@rutgers.edu \\
    $^*$Lizhou Fan and Wenyue Hua contribute equally. 
}

\begin{document}
\maketitle
\begin{abstract}
Complex reasoning ability is one of the most important features of current Large Language Models (LLMs), which has also been leveraged to play an integral role in complex decision-making tasks. Therefore, the investigation into the reasoning capabilities of LLMs is critical: numerous benchmarks have been established to assess the reasoning abilities of LLMs. However, current benchmarks are inadequate in offering a rigorous evaluation of the full extent of reasoning abilities that LLMs are capable of achieving. They are also prone to the risk of overfitting, as these benchmarks, being publicly accessible and static, allow models to potentially tailor their responses to specific benchmark metrics, thereby inflating their performance. 
Addressing these limitations, our research introduces a new benchmark, named \textbf{NPHardEval}. This benchmark is designed to evaluate the reasoning abilities of LLMs across a broad spectrum of 900 algorithmic questions, extending up to the NP-Hard complexity class. These questions are meticulously chosen to represent a wide range of complexity class below the NP-hard complexity class, offering a rigorous measure of the reasoning ability of LLMs. 
Through this study, we shed light on the current state of reasoning in LLMs, providing an objective and rigorous perspective through the comparison of LLMs' performance across complex classes. 
Our findings contribute significantly to understanding the current capabilities of LLMs in reasoning tasks and lay the groundwork for future advancements in enhancing the reasoning abilities of these models. 
Moreover, this benchmark is designed with a \textit{dynamic update mechanism}, where the datapoints are refreshed on a monthly basis. Such regular updates play a crucial role in mitigating the risk of LLMs overfitting to the benchmark, promoting a more accurate and reliable assessment of their reasoning capabilities. The benchmark dataset and code of \textbf{NPHardEval} are available at \url{https://github.com/casmlab/NPHardEval}.

\end{abstract}


\section{Introduction}
The advancement of LLMs has ushered in a transformative era in Artificial Intelligence (AI) research \cite{fan2023bibliometric}. One major advantage, believed by many researchers, is the unparalleled reasoning capabilities showcased by these models \cite{zhao2023survey}. Despite the implementation of various benchmarks for evaluating reasoning ability \cite{cobbe2021training, valmeekam2022large, chen2023theoremqa, hendrycks2020measuring, hendrycksmath2021}, existing methods reveal certain limitations. These include inadequacies in the precise characterization of reasoning abilities, the risk of models overfitting to specific benchmarks \cite{schaeffer2023pretraining}, and in some cases, the dependency on manual evaluation methods \cite{frieder2023mathematical}. Additionally, it is theoretically interesting to examine the extent to which LLMs can address problems in the computational complexity hierarchy \cite{johnson1990catalog}, especially NP-hard or NP-complete problems. In response to these issues and questions, we introduce a new benchmark \textbf{NPHardEval}, which leverages the well-established principles of computational complexity classes to provide a more rigorous and quantitative assessment of the reasoning abilities of large language models.

Our benchmark, meticulously designed to evaluate the reasoning abilities of Large Language Models (LLMs), comprises 9 carefully chosen reasoning tasks. These tasks are segmented according to complexity classes as outlined in \cite{johnson1990catalog}, with each class containing 100 instances distributed across 10 distinct levels of difficulty. This structured approach allows for a thorough and quantifiable assessment of LLMs' reasoning capacities. The selection of problems within our benchmark is particularly significant as it also mirrors the nuance of real-world decision-making and optimization scenarios, including critical areas such as logistics, scheduling, network design, and various other domains where optimal solutions carry substantial economic and practical implications. As LLMs are increasingly utilized in complex problem-solving scenarios, the need for an accurate and rigorous assessment of their reasoning abilities becomes paramount. Such an evaluation serves as a crucial and reliable indicator of their capabilities, guiding their effective integration and application in various contexts.
We can also gain deeper insights into both the strengths and limitations of their computational reasoning abilities.

Another novel feature of our benchmark is its end-to-end automation, encompassing both the generation of tasks and the verification of results. This automation is facilitated by the use of well-known tasks within the benchmark, for which mature and established algorithms have been developed to provide solutions. This systematic approach ensures a high degree of accuracy and reliability in the evaluation process, which also enables easy update of datapoints in the benchmark. This automated framework facilitates an effortless updating of datapoints within the benchmark. As a result, we design the benchmark to refresh its datapoints monthly, effectively reducing the likelihood of the model overfitting the dataset. This dynamic updating mechanism is crucial in maintaining the rigor and relevance of the benchmark over time. We welcome open submissions of model performance to our benchmark directly through the \textbf{NPHardEval} leaderboard on HuggingFace: \url{https://huggingface.co/spaces/NPHardEval/NPHardEval-leaderboard}.

In general, our benchmark offers several advantages compared with current benchmarks:
\begin{itemize}
    \item The questions in the benchmark utilized are grounded in the established computational complexity hierarchy, a concept extensively studied in theoretical computer science. This foundation enables us to leverage existing research to rigorously and quantitatively measure an LLM's logical reasoning extent.
    \item We incorporate automatic checking mechanisms for these questions, as they are based on algorithmically computable problems. Human intervention is not required to determine the correctness of the LLM's responses.
    \item The method allows for the automatic generation of questions so that we can update the benchmark on a monthly basis. This monthly-refreshed benchmark helps prevent model's overfitting as we can always generate novel questions with varying difficulty levels for evaluation. 
    \item The benchmark excludes numerical computation from the questions, which is notably difficult for LLM. This focus allows for a more accurate evaluation of an LLM's pure logical reasoning ability, as numerical computation can obscure this assessment.
    \item Our methodology offers insights into a long-standing and intriguing question within the field: the degree to which LLMs are capable of tackling problems classified as NP-hard or NP-complete.
\end{itemize}

In our research utilizing the benchmark, we aim to address three critical aspects to evaluate and understand the reasoning abilities of LLMs (Foundation Models):
\begin{enumerate}
    \item \textbf{Model Performance Comparison:} Our benchmark compares the reasoning ability of 12 closed-source models (including as GPT 4 Turbo, Claude 2, GPT 3.5 Turbo, Claude Instant, and PaLM 2) and open-source models (including Yi-34b, Qwen-14b, Mistral-7b, Phi-2, MPT-30b, Vicuna-13b, and Phi-1.5) across three complexity classes (P, NP-complete, and NP-hard) and 10 difficulty levels. This comparison can shed light on the relative strengths and weaknesses of these models and determine the proficiency of them in solving progressively challenging problems, thereby gauging their capability to handle tasks with escalating complexity.
    \item \textbf{Robustness of Benchmark Assessments:} This study examines whether the frequent updating of algorithmic benchmarks can effectively prevent the risk of ``hacking'' the benchmark. The dynamic updating of benchmarks is proposed as a strategy to reduce the likelihood of LLMs overfitting to these benchmarks. However, a pertinent question arises: does finetuning LLMs on benchmarks from the previous month lead to overfitting specific problem types? To explore this, we conducted an experiment where three open-source models -— Phi-2, Mistral-7b, and Qwen-14b —- were finetuned on five distinct versions of the benchmarks. The performance of these models was evaluated on two versions of the benchmark, each differing in difficulty level. This approach allowed us to assess whether finetuning enables models to ``hack'' benchmarks of varying complexity.
    \item \textbf{Generalization through In-context Learning:} Given examples in the context, can LLMs genuinely learn and apply algorithmic skills presented in contextual examples as opposed to merely mimicking problem-solving processes \cite{wei2023larger, min2022rethinking}? We differentiate between ``learning'' and ``mimicking'' by evaluating whether LLMs can generalize solutions to new problems of varying difficulty levels within the same task, after being exposed to examples. Our hypothesis is that if an LLM has truly learned the underlying algorithmic skill, it should be able to tackle problems across different difficulty levels within the same task. Conversely, if an LLM is merely mimicking, its performance may falter when faced with variations in problem difficulty.
\end{enumerate}

The contribution of this paper is as follows: We present the first complexity classes-based reasoning benchmark, \textbf{NPHardEval}. This benchmark enables a rigorous evaluation of LLMs' reasoning capabilities on a wide range of complex reasoning tasks with varying difficulty. Through the above research questions, we aim to provide a comprehensive analysis of the reasoning capabilities of LLMs, exploring their potential for genuine understanding and application of complex problem-solving skills. 

\begin{figure} 
    \centering
    \includegraphics[scale=.5]{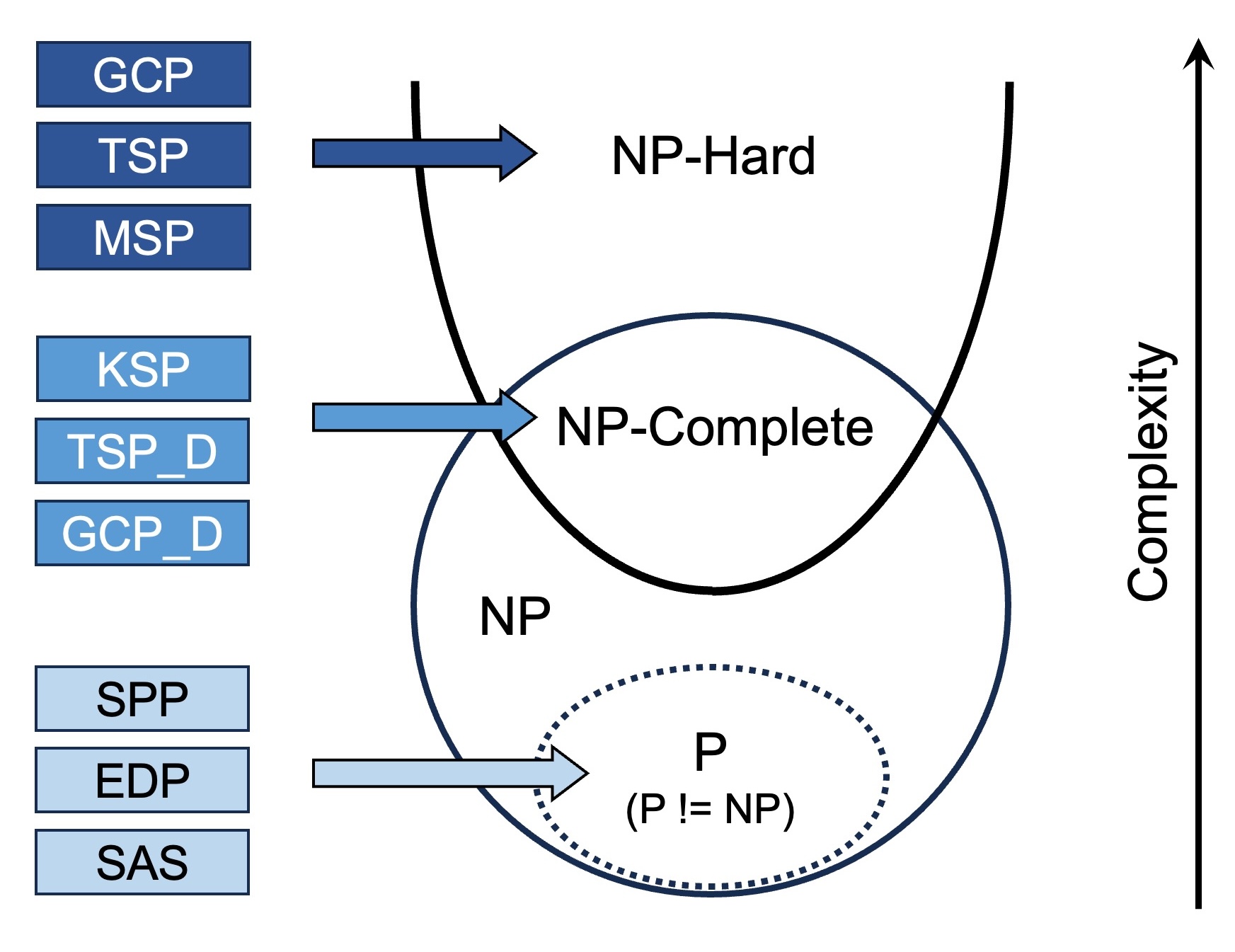}
    \caption{Computational complexity classes P, NP-complete, and NP-hard and corresponding tasks}
    \label{Fig:pnp_rel}
\end{figure}

\section{Related Work}

\subsection{Reasoning ability of LLMs}

\par
LLMs \cite{brown2020language, chowdhery2023palm, chung2022scaling} have made significant advancements in natural language processing and related fields. Recent research underscores the unprecedented reasoning abilities of LLMs in various fields, from biomedical and human-computer interaction research to humanities and social studies \cite{huang2022towards, hua2023war, fan2023datachat, gao2023examining, li2023hot}. It has been discussed that these models exhibit ``emergent'' behaviors, including the ability to ``reason'' when they are large enough \cite{wei2022emergent, schaeffer2023emergent}. By providing the models with the chain of thoughts with a simple prompt ``Let's think step by step'', these models are able to answer questions with explicit reasoning steps \cite{wei2022chain}. This has sparked considerable interest in the community since reasoning ability is a hallmark of human intelligence. Various variations of chain-of-thought have been developed to prompt models' reasoning ability\cite{kojima2022large, wang2022iteratively, hua2022system}, such as tree of thought\cite{yao2023tree}, graph of thought\cite{besta2023graph}, self-inspring technique\cite{wang2023recmind}.

Later, various self-critique methods have been proposed to enhance LLM's reasoning performance. The Recursively Criticizes and Improves (RCI) approach, for example, iteratively refines outputs, proving more effective in automating computer tasks and elevating reasoning capabilities \cite{kim2023language}. Similarly, backward verification proposes an intuitive human-like mechanism for LLMs to self-check and improve their conclusions, reducing errors in reasoning tasks \cite{weng2022large}. Moreover, the interplay of reasoning and action showcases LLMs' outstanding synergistic ability. For instance, the ``ReAct'' approach highlights that reasoning can enhance action plans, while actions can help the model interface with external sources for better reasoning \cite{yao2022react}. In addition, the capability of LLMs to learn from feedback also indicates their reasoning potential. ``Reflexion'' is a workflow that reinforces LLMs through linguistic feedback without updating model weights \cite{shinn2023reflexion}. In robotic contexts, LLMs demonstrate enhanced performance with environment feedback, creating an internal ``monologue'' to assist decision-making \cite{huang2022inner}. 

Despite the impressive performance exhibited by LLMs, there remains a gap in our rigorous understanding of the extent and depth of reasoning these models are capable of. Our paper aims to address this gap by providing a framework to study the reasoning abilities of LLMs within the well-established hierarchy of computational complexity. This approach seeks to systematically evaluate and quantify their reasoning capabilities in a more structured and academically rigorous manner.


\subsection{Benchmarks of LLMs' Performance}

\par The advancement of LLMs has catalyzed the evolution of a range of general-purpose AI technologies, underscoring the importance of accurately assessing these models' reasoning capabilities. Existing evaluation approaches predominantly rely on datasets comprising human-generated questions and their standard answers. For instance, MMLU \cite{hendrycks2020measuring} and GAOKAO \cite{zhang2023evaluating} both utilize human exam questions in their automated evaluations. Additionally, datasets such as the French National Math Exam, Hungarian National High School Exam\footnote{https://huggingface.co/datasets/keirp/hungarian\_national\_hs\_finals\_exam}, and GHOST (Graduate-Level High-Order Skill Tests)\cite{frieder2023mathematical} are utilized to assess LLMs' reasoning proficiency. \cite{zhu2023dyval} proposes a dynamic graph-based reasoning benchmark. These sources aim to ensure the absence of data leakage. Nonetheless, the requirement for manual verification of answers in these datasets limits their practical utility. 
In general, these datasets are commonly employed as benchmarks in the field; however, they lack a quantitative metric for assessing the difficulty level of the questions and the extent of reasoning necessary to answer them. This absence of precise measurement criteria results in a limited understanding of the logical reasoning capabilities of large language models.

\par Other Benchmarks such as AlpacaEval \cite{dubois2023alpacafarm} and SuperCLUE \cite{xu2023superclue} have attempted to incorporate open-ended questions in English and Chinese, respectively, to capture a diverse breadth of possible answers and enhance the comprehensiveness of LLM's evaluation. However, they are not universal and are often constrained by language barriers and cultural contexts, potentially skewing the evaluation of reasoning abilities toward a specific scenario. Reasoning tasks should transcend linguistic and cultural specifics, focusing instead on universal logical principles.
Big-Bench Hard \cite{suzgun2022challenging}, DROP \cite{dua2019drop}, and HellaSwag \cite{zellers2019hellaswag}, while valuable, predominantly target multi-step reasoning, reading comprehension, and commonsense reasoning, respectively. They do not adequately prioritize complex logical reasoning in their assessment criteria. 

The prevalent focus on question answering and math problems in current benchmarks may insufficiently capture the essence of reasoning - the ability to logically process and deduce information beyond memorized knowledge. It also falls short on providing a rigorous metric on the reasoning ability. This gap highlights the need for a paradigm expansion in LLM evaluation, calling for logic-based reasoning benchmarks to complete the traditional utility-based approach, where we have quantitative evaluation on the computational complexity of the questions, indicating the amount of reasoning ability required.  

\section{Benchmark Construction}

\subsection{Complexity Classes}

In our study, we employ the concept of complexity classes to categorize the reasoning tasks for LLMs. These classes are defined based on the computational resources, such as time or memory, required to solve the problems they contain \cite{johnson1990catalog}. Primarily, most complexity classes comprise decision problems that can be solved using a Turing machine, with differentiation based on their time or space (memory) requirements.
For example, the class P includes decision problems that a deterministic Turing machine can solve in polynomial time. Tasks within this class often pose multi-dimensional cognitive challenges, enriching the evaluation framework of LLMs. This structured approach not only aids in assessing the reasoning capabilities of LLMs but also holds substantial relevance in various practical applications, particularly in optimization and high-level decision-making scenarios.

In particular, we use three complexity classes to define the task complexity in the benchmark, including P (polynomial time), NP-complete (nondeterministic polynomial-time complete), and NP-hard, which are increasingly complex in both the intrinsic difficulty and the resources needed to solve them. Figure \ref{Fig:pnp_rel} shows their relation regarding computational complexity in an Euler diagram. 
This approach aims to delineate the extent of complex reasoning achievable by LLMs, thus for each complexity class, we only choose tasks from the non-overlapping subset of the complexity class. In our selection criteria, we intentionally exclude tasks that demand intensive mathematical computations, such as matrix multiplication and logarithmic calculations. Thus, we do not list NP class (questions in NP but not P and not NP-complete), which is exemplified by the discrete logarithm and integer factorization problems, as the majority of such problems are characterized by their calculation-intensive nature (see details in Appendix \ref{Appnedix:task_exclude}).

\subsubsection{P (Polynomial time) Tasks}


This class consists of tasks that can be solved by a deterministic Turing machine in polynomial time. Essentially, it represents tasks that are efficiently solvable. We include three P problems in the benchmark, namely Sorted Array Search (SAS), Edit Distance Problem (EDP), and Shortest Path Problem (SPP). 


\paragraph{Sorted Array Search (SAS)}
SAS is about finding the position of a target value after sorting a given array. Given an array $A$ of $n$ elements and a target value $T$, the goal is to determine the index at which $T$ is located in $A$ after sorting. Renowned algorithms like binary search efficiently accomplish this task by iteratively halving the search interval, operating in logarithmic time.
The problem can be formally stated as finding an index $i$ such that $A[i] = T$, or determining that no such index exists. It is commonly used in databases and search engines to quickly find specific data within a large dataset \cite{kipf2019sosd}. 

\paragraph{Edit Distance Problem (EDP)}
EDP is about finding the minimum number of operations required to transform one string into another. Given two strings, $A$ and $B$, of lengths $m$ and $n$ respectively, the aim is to determine the minimum number of operations needed to convert $A$ into $B$. The allowable operations are insertion, deletion, and substitution of a single character. Formally, the problem can be defined as finding a minimum number $d$ such that string $A$ can be transformed into string $B$ using $d$ operations. 
This algorithm has a time complexity of $\mathcal{O}(ab)$ where $a$ and $b$ are the lengths of the strings. When the full dynamic programming table is constructed, its space complexity is also $\mathcal{O}(ab)$.
EDP has widespread applications, especially in fields like computational biology for sequence alignment, natural language processing for spell checking and correction, and in data analysis for measuring similarity between data strings.

\paragraph{Shortest Path Problem (SPP)}
SPP is about finding the shortest path between two nodes in a non-negative weighted graph. In our experiments, we ask for the shortest path between the first and last nodes. Given a graph $G = (V, E)$ with a weight function $w: E \rightarrow \mathbb{R}$ assigning weights to edges, and two vertices $u$ and $v$ in $V$, the task is to find the path from $u$ to $v$ that minimizes the total weight. This is often solved using Dijkstra's algorithm which systematically expands the shortest path from the starting node until it reaches the target node. Formally, the problem is to find a path $P = (v_1, v_2, ..., v_k)$, where $v_1 = u$ and $v_k = v$, such that the sum of weights of consecutive edges in $P$, $\sum_{i=1}^{k-1} w(v_i, v_{i+1})$, is minimized. This problem can be used in network routing, GPS navigation systems, and logistics to find the shortest or most efficient path between two points. It helps in reducing travel time and costs in transportation and communication networks. 

\subsubsection{NP-complete problems}

This is a subset of NP. A problem is NP-complete if it is in NP and as hard as any problem in NP. If any NP-complete problem can be solved in polynomial time, then every problem in NP can also be solved in polynomial time. We include three NP-complete problems that are not in P in the benchmark, namely Traveling Salesman Problem Decision Version (TSP-D), Graph Coloring Problem Decision Version (GCP-D), and Knapsack Problem (KSP).

\paragraph{Traveling Salesman Problem (Decision Version, TSP-D)}
TSP-D is concerned with determining if a salesman can complete a route, visiting each city at least once, with the total travel distance being less than a specified value. Given a complete graph $G = (V, E)$ with vertices $V$ representing cities and edges $E$ representing paths between cities, each edge $(i, j)$ is assigned a distance $d(i, j)$. The decision version of this problem asks whether there exists a tour (a sequence of cities) such that the total distance of the tour is less than or equal to a given value $D$. Formally, the problem can be stated as finding a permutation $P$ of the set of cities ${1, 2, ..., n}$ that satisfies the condition $\sum_{i=1}^{n-1} d(P(i), P(i+1)) + d(P(n), P(1)) \leq D$. This problem is useful in logistics and supply chain management in planning efficient delivery routes and schedules \cite{roberti2021exact}. 

\paragraph{Graph Coloring Problem (Decision Version, GCP-D)}
GCP-D involves determining if it is possible to color the vertices of a graph using a given number of colors so that no two adjacent vertices share the same color. Given an undirected graph $G = (V, E)$, with $V$ representing vertices and $E$ representing edges, the goal is to find out if there is a way to assign one of $k$ colors to each vertex such that for any edge $(u, v) \in E$, the vertices $u$ and $v$ have different colors. The formal statement is to determine if there exists a coloring function $c: V \to {1, 2, ..., k}$ such that for every edge $(u, v) \in E$, $c(u) \neq c(v)$. It has wide applications in Round-Robin Sports Scheduling, Aircraft scheduling, and Biprocessor tasks \cite{ahmed2012applications}. 

\paragraph{Knapsack Problem (KSP)}
KSP asks whether a subset of items can be chosen to fit into a knapsack of fixed capacity without exceeding it, while also maximizing the total value of the selected items. Consider a set of items, each with a weight $w_i$ and a value $v_i$, and a knapsack with a weight capacity $W$. The problem is to select a subset of these items such that the total weight does not exceed $W$ and the total value is maximized. Formally, let $x_i$ be a binary variable indicating whether item $i$ is included in the knapsack ($x_i = 1$) or not ($x_i = 0$). The problem can be stated as maximizing $\sum_{i=1}^{n} v_i x_i$ subject to the constraint $\sum_{i=1}^{n} w_i x_i \leq W$, where $n$ is the number of items. It is used in resource allocation and budgeting where the goal is to maximize the total value of a selection under a weight or cost constraint. Applications include cargo loading, and electric vehicle charging \cite{sun2020competitive, cho2019knapsack}.



\subsubsection{NP-hard problems}

These problems are at least as hard as the hardest problems in NP. They may not necessarily be in NP (i.e., they may not have solutions verifiable in polynomial time) but solving an NP-hard problem in polynomial time would imply that P = NP. We include three NP-hard problems that are not reducible to NP-complete problems in the benchmark, namely Traveling Salesman Problem Optimization Version (TSP), Graph Coloring Problem Optimization Version (GCP), and Meeting Scheduling Problem (MSP).

\paragraph{Traveling Salesman Problem (Optimization Version, TSP)}
TSP-O involves finding the shortest route for a salesman to visit each city exactly once and return to the starting city. Given a complete graph $K_n$ with $n$ vertices, where each vertex represents a city and each edge $(i, j)$ is assigned a non-negative cost or distance $d(i, j)$, the problem is to find the shortest possible route that visits each city exactly once and returns to the origin city. Formally, let $P$ be a permutation of the set of cities ${1, 2, ..., n}$ representing the order in which the cities are visited. The traveling salesman problem can be formulated as finding the permutation $P$ that minimizes the total travel cost, given by the function $f(P) = d(P(n), P(1)) + \sum_{i=1}^{n-1} d(P(i), P(i+1))$. This problem is important in operational research and logistics to find the most efficient route to visit multiple locations and return to the origin, particularly route planning for delivery services, maintenance operations, and sales. 

\paragraph{Graph Coloring Problem (Optimization Version, GCP)}
GCP-O refers to the problem of coloring vertices of a graph in such a way that no two adjacent vertices have the same color. Given an undirected graph $G = (V, E)$, where $V$ is the set of vertices and $E$ is the set of edges, assign a color to each vertex such that no two adjacent vertices have the same color. Formally, let $c: V \to C$ be a function that assigns a color from a set of colors $C$ to each vertex in $V$. The graph coloring problem can be formulated as finding a proper coloring, i.e., a function $c$ such that for every edge $(u, v) \in E$, $c(u) \neq c(v)$. This problem is used in constraint satisfaction problems and applied in exam timetabling and register allocation in compilers \cite{lintzmayer2011register}. 

\paragraph{Meeting Scheduling Problem (MSP)} 
MSP deals with allocating time slots for meetings such that all constraints, including participant availability and room capacity, are satisfied without overlaps. Given a set of $n$ participants and their availability for $m$ time slots, find a schedule that maximizes the number of participants who can attend the meeting. Formally, let $A = {a_1, a_2, ..., a_n}$ be the set of participants and $T = {t_1, t_2, ..., t_m}$ be the set of time slots. For each participant $a_i$, let $S_i$ be a subset of $T$ representing the times when $a_i$ is available and $m_i$ be a subset of meetings that are required to attend. The meeting scheduling problem can be formulated as finding a subset $S \subseteq T$ such that $|{a_i \in A | S_i \cap S \neq \emptyset}|$ is maximized. In other words, the aim is to find a scheduling subset $S_i$ where the collective availability of participants intersects with $S_i$, ensuring maximum participation. This problem is crucial in organizational management for scheduling meetings involving multiple participants with varying availability. It ensures optimal utilization of time and resources and is used in corporate scheduling systems and collaborative software \cite{bofill2022constraint}. 


\subsection{Difficulty Level for Tasks}

\textbf{NPHardEval} categorizes the challenges it presents into a hierarchy of difficulty, spanning from the simplest to the most complex. This structure is divided into 10 distinct levels of difficulty for each task, with the initial level being designed as the most basic challenge that an LLM might face. This gradation allows for a nuanced assessment of an LLM's problem-solving abilities across a spectrum of increasingly complex tasks. For instance, the GCP-D problem has difficulty levels 1 to 10 with questions of 6, 8, 10, 12, 14, 16, 18, 20, 22, and 24 average edges and 6, 7, 8, 9, 10, 11, 12, 13, 14, and 15 nodes. Beginning with graphs of 6 nodes and 6 edges, each subsequent level incorporates an additional 2 edges and 1 node, culminating in graphs of  24 edges and 15 nodes at the most challenging level.

The difficulty level is not strictly bound to a linear scaling of difficulty; rather, it is designed to explore the nuances of performance degradation. By observing how LLMs cope with an escalating series of challenges, we aim to identify the inflection point where the performance notably diminishes. This approach provides a comprehensive understanding of where LLMs excel and where they falter, informing potential pathways for the enhancement of their reasoning capabilities.

\subsection{Data Synthesis}

In the context of data synthesis for complex tasks, the approach can be categorized into two distinct methodologies, each corresponding to a different type of data structure: graph data (e.g., GCP) and linear data (e.g., MSP). The synthesis process in both cases is governed by a progression of complexity across a spectrum of predefined levels. This structured approach enables the creation of diverse datasets, suitable for evaluating and benchmarking LLMs' reasoning ability. We provide examples of the synthesized data and how thay are used in prompts in Appendix \ref{Appendix:eg_experiment}.

\paragraph{Graph Data Synthesis} The complexity in graph data synthesis escalates through a series of levels, each defined by a set of parameters that dictate the graph's size and intricacy. These parameters typically include the number of vertices, the number of edges, and the range of edge weights. At lower levels, graphs are simpler with fewer vertices and edges, and a limited range of edge weights. As the level increases, the graphs become progressively more complex, featuring more vertices, a higher density of edges, and a wider variety of edge weights. The synthesis process is as follows:
\begin{itemize}
    \item A generative function is employed to construct individual graph instances. This function adheres to the principles of graph theory, ensuring the creation of simple graphs without self-loops and duplicate edges, and respecting the parameters dictated by the current difficulty level.
    \item A batch synthesis function then iteratively employs the generative function to produce multiple graph instances across the spectrum of difficulty levels.
    \item Finally, the synthesized graph instances are preserved in a tabulated format (in a CSV file), facilitating subsequent utilization and analysis.
\end{itemize}

\paragraph{Linear Data Synthesis} In linear data synthesis, complexity is modulated by manipulating the length of the data array and the range of its constituent elements. Initial levels are characterized by shorter arrays with elements drawn from a narrow range. As the difficulty level ascends, the arrays lengthen, and the range of possible element values expands, thus introducing greater variability and complexity to the problem. The synthesis process is as follows:
\begin{itemize}
    \item A linear data instance generation function is first utilized. This function produces sorted arrays of random numbers within a defined range, and selects a target number, ensuring its presence within the array to guarantee solvability.
    \item Multiple instances are generated through an iterative process, adhering to the difficulty levels outlined.
    \item These instances are then systematically recorded in a structured format (in a JSON file) for easy access and analysis.
\end{itemize}

\section{Experimental Setting}
This section presents the experiment setting to answer the three research questions. Our approach assessed 10 distinct LLMs, with a dichotomy between five proprietary (closed-source) models and five open-source models, including GPT 4 Turbo, Claude 2, GPT 3.5 Turbo, Claude Instant 1.2, PaLM 2,  Vicuna-13b, Yi-34b, Mistral-7b, MPT-30b, and Phi-1.5.

\subsection{Experiment 1: Model Performance Comparison}
To evaluate the reasoning abilities of different LLMs through the \textbf{NPHardEval} benchmark, we employed a comparative experimental design. We use zero-shot prompts as the foundational measure of performance. These prompts comprise a task description and a specific question, presented without any preceding examples, to gauge the base capability of the model. 

The complexity of the problems presented to the models spanned from polynomial-time (P) to NP-complete and NP-hard levels. To ensure comprehensive coverage, we utilize the full set of 900 problems in \textbf{NPHardEval}, capturing the multifaceted nature of real-world challenges that typically exceed the capabilities of straightforward algorithmic approaches. Each model's performance was evaluated based on two primary metrics: weighted accuracy and failure rate across the different complexity classes of problems, as we discussed in Section \ref{sec:eval}.

To evaluate across task complexity, specifically comparing the complexity among P, NP-Complete, and NP-Hard pairs, we initially pinned the data based on complexity levels. Subsequently, we applied the Wilcoxon test to each pair of complexity sets. Wilcoxon is a non-parametric statistical hypothesis test that allows us to compare two populations with matched samples. To evaluate problem difficulty, aiming to discern differences among problems within the complexity category, we pinned the data based on the specific problems and then used the Wilcoxon test to compare pairs of different problem sets. 

\subsection{Experiment 2: Benchmark Robustness}
The primary objective of this experiment is to ascertain whether it is possible to ``hack'' our benchmark by finetuning models on its previous versions. To simulate this, we constructed five versions of the benchmark, maintaining a consistent difficulty level. Additionally, we utilize two distinct versions of the benchmark, each varying in difficulty, to evaluate the potential for hacking under varying conditions. To replicate the progression of time, models were finetuned sequentially on one to five benchmarks, each finetuned checkpoint is tested on the two distinct benchmarks for evaluation.

The experiment involved finetuning three high-performing open-source models: Phi-2, Mistral-7b, and Qwen-14b. Due to constraints in computing resources, the Yi-34b model was not included in the finetuning process. For the finetuning process, we employed the QLoRA technique, applying specific hyperparameters: batch size set to 8, a single epoch, a warmup proportion of 0.03, a learning rate of 1e-4, lora\_r at 64, lora\_alpha at 16, and a lora\_dropout of 0.1. This approach aims to rigorously test the robustness of our benchmark against potential overfitting strategies. 

\subsection{Experiment 3: Comparative Analysis of Learnability by In-context Learning}
A prevalent approach in current few-shot learning involves using examples that bear similarity to the test question. However, this raises a question about the extent to which the model is replicating the problem-solving process from the examples as opposed to genuinely acquiring reasoning skills. Consequently, it becomes pertinent to investigate whether the problem-solving abilities developed through example-based learning are generalizable. 

To delve deeper into the models' in-context learning abilities, we utilize various few-shot in-context learning prompts to discern whether the model is ``learning'' from the few-shot examples or merely ``mimicking'' the behavior. In our benchmark, since we distinctly classify the difficulty level of each question, it allows for the use of questions from the same task but with varying difficulty levels as few-shot examples. The crux of this analysis lies in varying the difficulty levels of examples within the prompts. Since the fundamental algorithmic skill required to solve a question remains constant across varying difficulty levels under the same task, a model that truly learns this skill should show consistent performance irrespective of the example difficulty in the prompt. We propose the following hypotheses about the relationship between in-context learning ability and the difference of difficulty level between the given examples and the question being asked in context:

\begin{itemize}
\item Models possessing optimal generalization capabilities should demonstrate consistent performance improvement regardless of the difficulty level of the prompt examples in context. This assumption is based on the premise that models with robust learning abilities are capable of discerning and applying the intrinsic problem-solving skills learned in the examples. Given that questions within the same task fundamentally require similar skills, variations in difficulty are unlikely to significantly affect the model's performance. 
\item If a model exhibits the ability to generalize only from some types of examples but is unable to extend this learning to others, it reveals a deficiency in its capacity for generalization in terms of reasoning. This suggests that the model is not genuinely acquiring problem-solving skills from the examples but merely recognizing and applying patterns from examples that are of equal or greater complexity to the problem at hand. 
\item If a model is unable to generalize from either more difficult or easier examples and is restricted to examples of the same difficulty level, it strongly suggests that the model is merely replicating the process presented in the context rather than internalizing any fundamental problem-solving techniques or pattern recognition embedded within the examples. This behavior indicates a profound deficiency in the model's ability to comprehend and understand the underlying principles. It points to an absence of transferable, logic-learning skills, reflecting a superficial form of learning that is limited to surface-level imitation rather than a deeper, conceptual grasp.
\end{itemize}

We categorize the few-shot prompts into three types:

\begin{itemize}
    \item Few-shot prompts with examples of the same difficulty level: Here, the model is provided with five examples in the prompt, all of which are at the same difficulty level and distinct from the question being asked.
    \item Few-shot prompts with examples that are easier than the question: This set comprises five variations of prompts, each with examples that are 1, 2, 3, 4, and 5 levels easier than the question, respectively.
    \item Few-shot prompts with examples that are more challenging than the question: Similarly, we prepare five sets of prompts, each containing examples that are 1, 2, 3, 4, and 5 levels more difficult than the question, offering a gradient of increased challenge.
\end{itemize}

Through this diverse array of prompts, we aim to provide a nuanced understanding of the LLMs' ability to learn from examples, thereby offering valuable insights into their underlying learning capabilities.

\subsection{Evaluation Metrics}
\label{sec:eval}

To evaluate the reasoning ability of LLMs, we utilize two metrics, the Weighted Accuracy and the Failure Rate, to comprehensively quantify the correctness of LLMs' reasoning outputs. 

\paragraph{Weighted Accuracy (WA)} is calculated for each problem either through the comparison with the correct answer or through step-by-step results checking, for those problems without the only answer. To better represent the comparative accuracy, we assign weights to different difficulty levels so that each level has a weight corresponding to its relative importance or challenge. Higher difficulty levels are given more weight in a linear manner (e.g., level 1 has weight 1, level 2 has weight 2, etc.). The Weighted Accuracy is formally defined as:
$$ WA = \frac{\sum_{i=1}^{10} (w_i \times A_i)}{\sum_{i=1}^{10} w_i} $$
where $w_i$ represents the weight assigned to difficulty level $i$, from 1 to 10, and $A_i$ is the accuracy at that level.

\paragraph{Failure Rate (FR)} is a measure used to assess the frequency of unsuccessful outcomes across the different problems and difficulty levels. It is particularly useful for identifying cases where an LLM's result does not comply with the expected output format. The Failure Rate is calculated by considering the proportion of failed attempts relative to the total number of attempts for each difficulty level. An attempt is defined as failed if the model generates results that cannot be successfully parsed in all endpoint calls, and we set the maximum times of try as 10. For each problem, the Failure Rate is then aggregated across all difficulty levels, taking into account the total 10 attempts at each level. The formal definition of Failure Rate is given by:
$$ FR = \frac{\sum_{i=1}^{10} F_i}{100} $$
where \( F_i \) denotes the number of failed attempts at difficulty level \( i \).


\section{Results}

\subsection{Reasoning Ability of Foundation Models}

\begin{figure}[ht] 
    \centering
    \includegraphics[width = \textwidth]{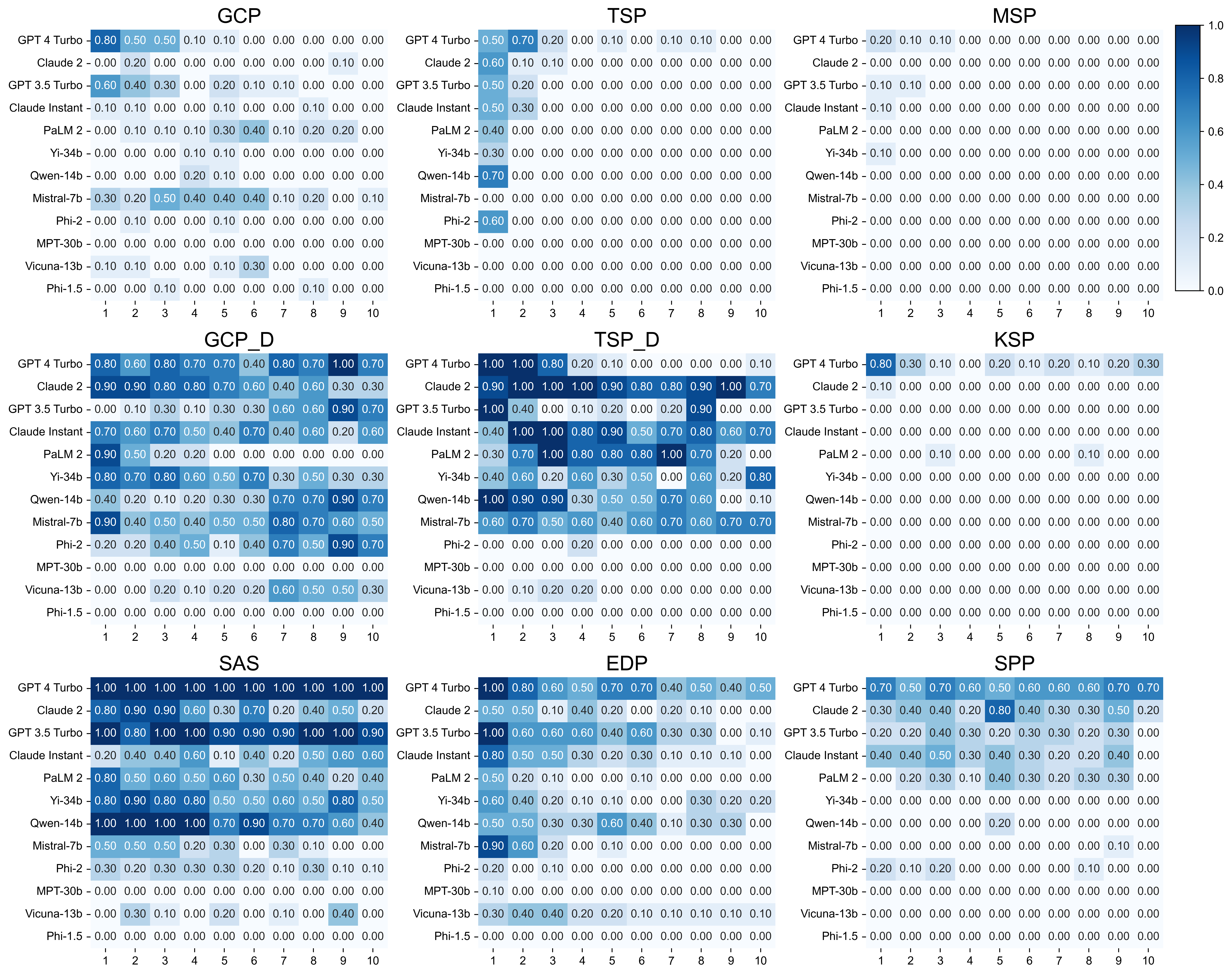}
    \caption{Zero-shot model performance on the nine tasks from P to NP-Complete bottom-up.}
    \label{Fig:zeroshot_heatmap}
\end{figure}

Experiment 1 focuses on a comprehensive comparison among various foundation models and across complexity classes and difficulty levels. In Figure \ref{Fig:zeroshot_heatmap}, we present the overall zero-shot accuracy for each problem, providing a visual representation of the performance of different models. 

Our observations reveal that closed-source models generally demonstrate higher accuracy and a lower rate of failure compared to their open-source counterparts. Notably, GPT-4 Turbo often emerges as the frontrunner in performance across the majority of tasks, indicating its superior problem-solving capabilities, while Claude 2, on the other hand, often performs the best on medium-level (NP-complete) complexity in zero-shot settings.

Within the realm of open-source models, Yi-34b, Qwen-14b, and Mistral-7b distinguish themselves by significantly outperforming other models in this category. We observe a disparity between the performance of these three models and other open-source options, highlighting a notable performance gap and suggesting that these models possess more advanced reasoning abilitie.

In particular, we use the weighted accuracy and the failure rate metrics to further quantify different models' performance. The trends observed below in both weighted accuracy and failure rates point to a nuanced understanding of the capabilities and limitations of current LLMs. We also utilize performance comparison tests within and across complexity classes, to further explore the model performance differences among complexity classes.

\begin{figure} 
    \centering
    \includegraphics[width = 0.9 \textwidth]{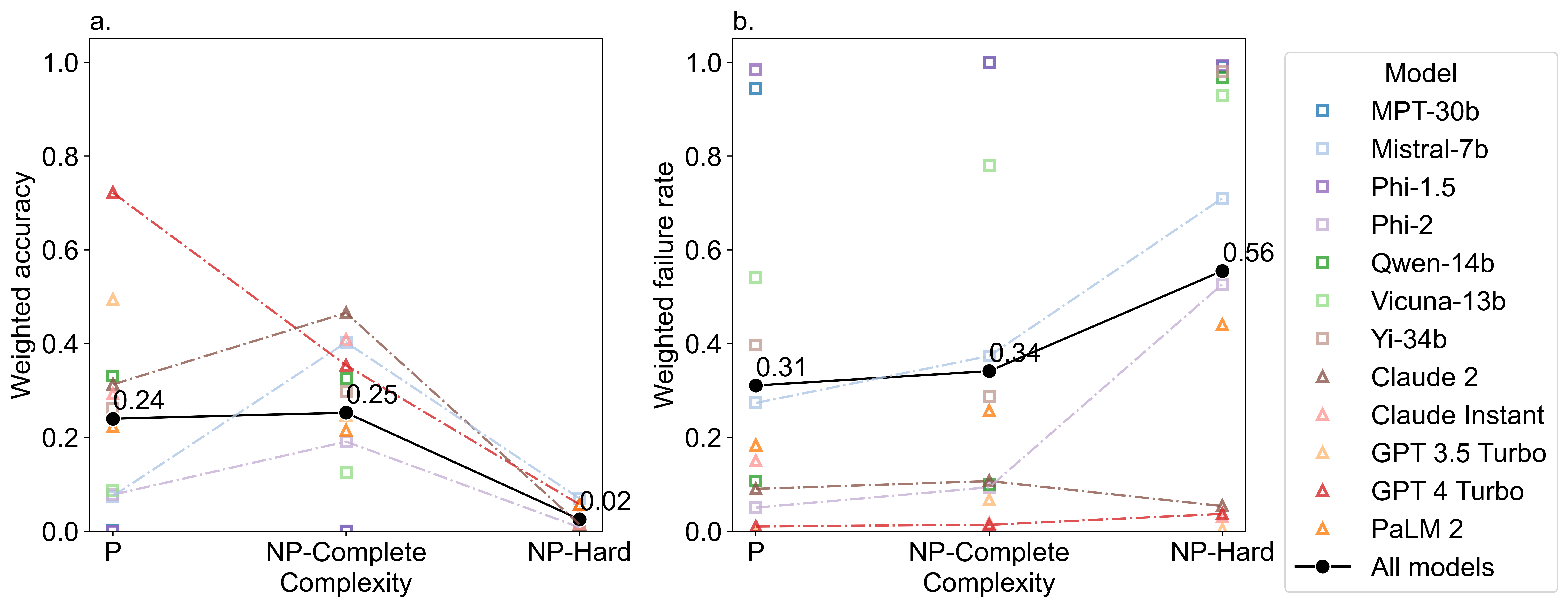}
    \caption{Model performance on different complexity problems: (a) weighted accuracy (b) (weighted) failure rate. Open models are denoted in squares and close models are denoted in triangles. Trends of metrics are demonstrated for models with outstanding performances in both weighted accuracy and failure rate, including both close-source (GPT 4 Turbo and Claude 2) and open-source (Mistral-7B and Phi-2) models.}
    \label{Fig:weighted_avg_acc}
\end{figure}

\paragraph{Weighted Accuracy} As Figure \ref{Fig:weighted_avg_acc}(a) shows, upon analysis of the weighted accuracy for different models across problem complexities, we observed a general trend where all models experienced a decrease in accuracy as problem complexity increased. Notably, there are two detailed findings for overall reasoning ability change. First, regarding the performance decay speed, among the 12 models we tested, the average performance demonstrated a higher accuracy at the P and NP-Complete complexity levels (with similar weighted accuracies of 0.24 and 0.25) but saw a sharper decline as the problems became more complex when proceeding to the NP-hard level (with a weighted accuracy of 0.02). There is a \textit{performance decay} on average when models are tested against NP-Hard problems. Second, close-source models usually perform better than open-source models -- there are more triangles in the upper locations than squares in Figure \ref{Fig:weighted_avg_acc}(a). 

\paragraph{Failure Rate} As Figure \ref{Fig:weighted_avg_acc}(b) indicates, the failure rates mirrored the trends observed in weighted accuracy but in reverse. On average, the models showed an increase in failure rates corresponding to the complexity of the problems. Similar to the weighted accuracy, open-source models fail more often (with more squares on the top) than the close-source models (with more triangles on the bottom), indicating close-source's models advanced ability in following the prompt to understand the reasoning problems and generate answers with correct format. 

\subsubsection{Performance across Task Complexity and Difficulty Levels}
\begin{figure}[ht] 
    \centering
    \includegraphics[width=\textwidth]{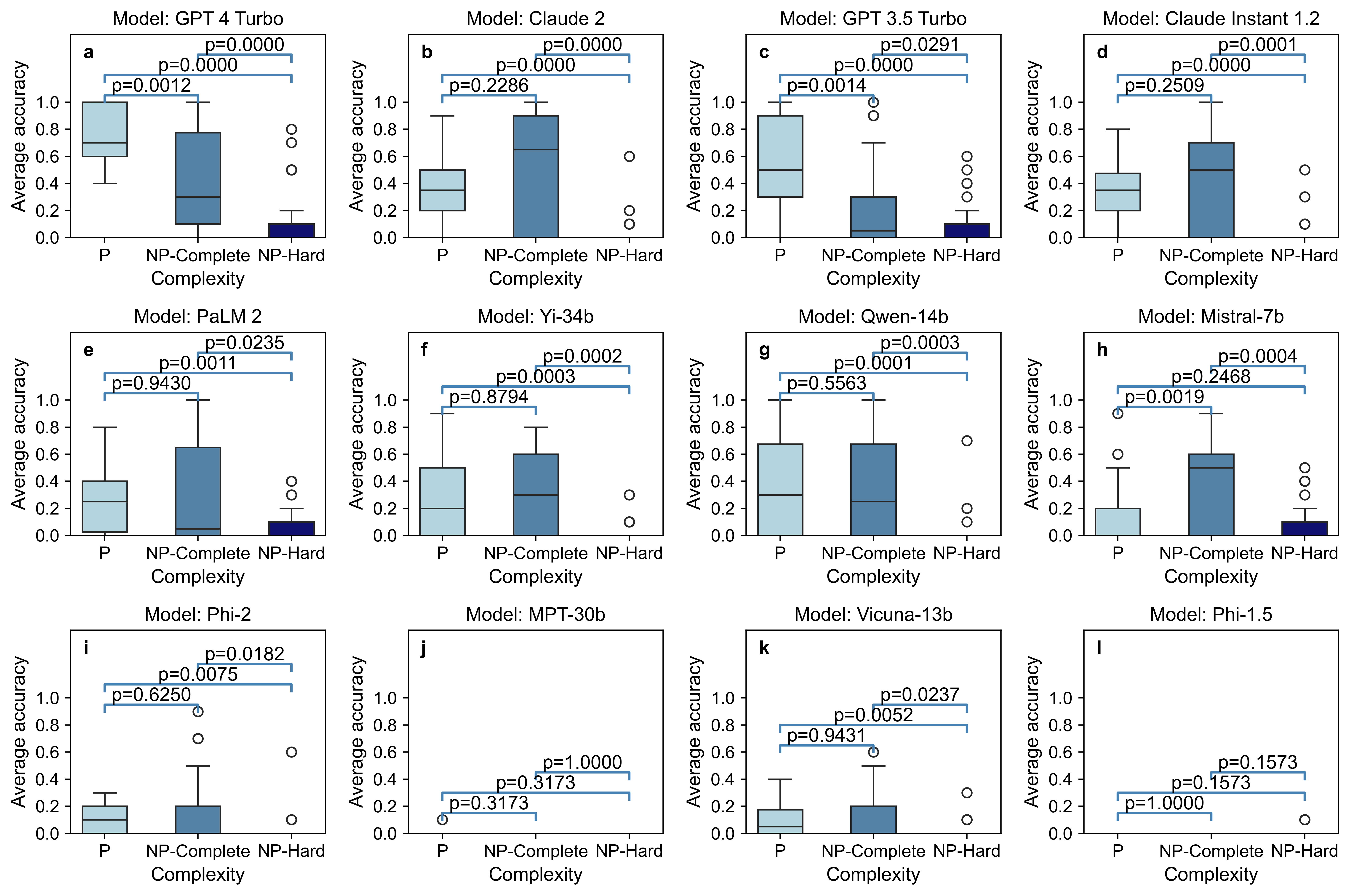}
    \caption{Models' performance on each complexity level. (a) GPT 4 Turbo. (b) Claude 2. (c) GPT 3.5 Turbo. (d) Claude Instant 1.2. (e) PaLM 2. (f) Yi-34b. (g) Qwen-14b. (h) Mistral-7b. (i) Phi-2. (j) MPT-30b. (k) Vicuna-13b. (l) Phi-1.5.}
    \label{Fig:complexities}
\end{figure}

Figure \ref{Fig:complexities} shows the accuracy of each model across different complexity levels. The test results reveal statistical significance ($p<0.05$) in the p-values between P and NP-Hard, as well as NP-Complete and NP-Hard. These findings indicate that our investigated LLMs performed significantly worse when confronted with NP-Hard problems compared to P and NP-Complete problems. 


\begin{figure}[ht] 
    \centering
    \includegraphics[width=\textwidth]{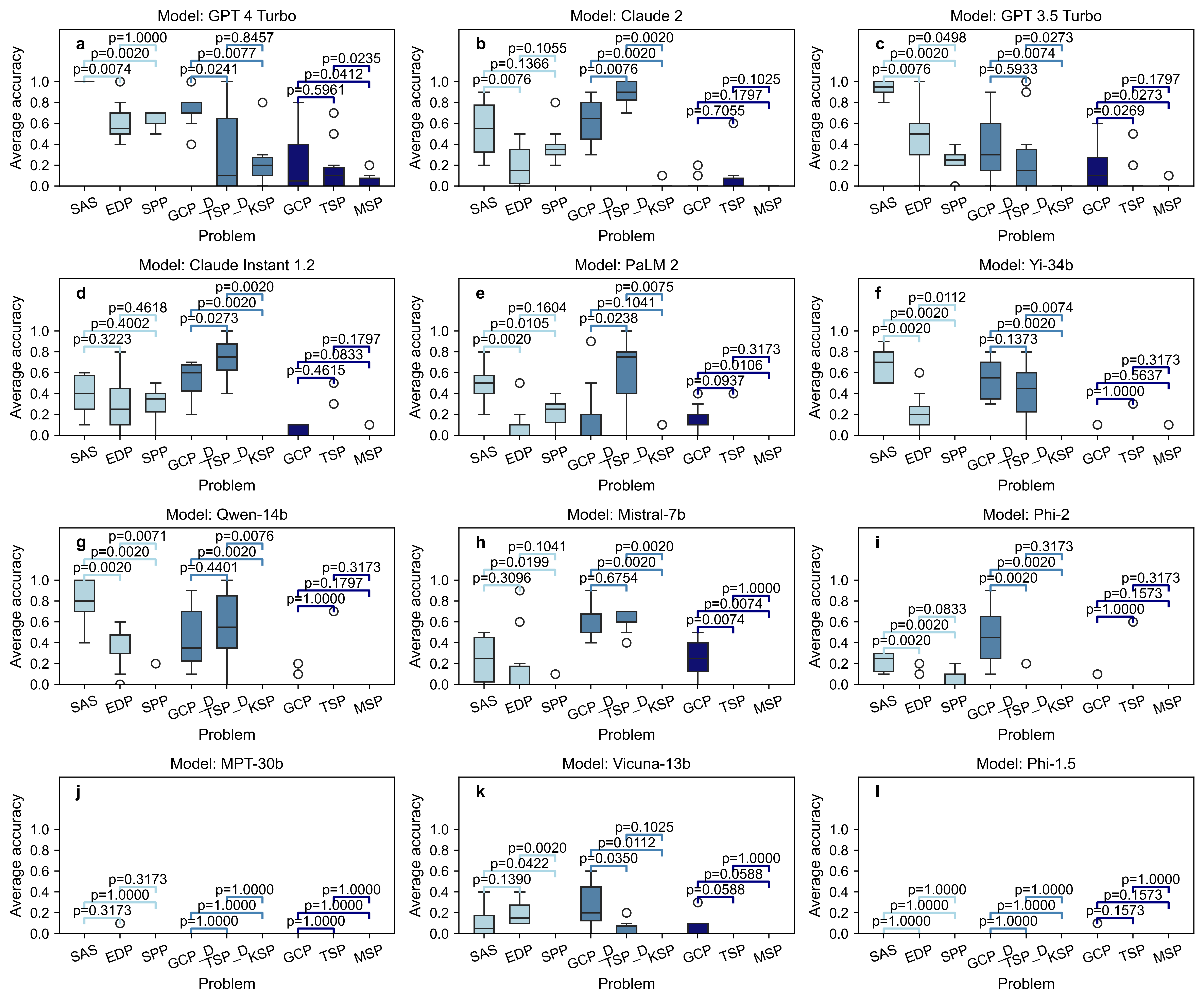}
    \caption{Models' performance on tasks across complexity levels. (a) GPT 4 Turbo. (b) Claude 2. (c) GPT 3.5 Turbo. (d) Claude Instant 1.2. (e) PaLM 2. (f) Yi-34b. (g) Qwen-14b. (h) Mistral-7b. (i) Phi-2. (j) MPT-30b. (k) Vicuna-13b. (l) Phi-1.5.}
    \label{Fig:problems}
\end{figure}

Figure \ref{Fig:problems} presents the accuracy of each model across various problems associated with P, NP-Complete, and NP-Hard complexities. Regarding P complexity, notable differences emerged among the models. GPT 3.5 Turbo, GPT 4 Turbo, Yi-34b, and Qwen-14b models exhibited significantly superior performance on the SAS problem compared to the other two problems. GPT 3.5 Turbo, Yi-34b, and Vicuna-13b models demonstrated markedly better performance on the EDP problem compared to the SPP problem. Only the Vicuna-13b model displayed slightly better performance, although not significant, on the EDP problem compared to SAS across all investigated models. 

Other observations include: GPT 4 Turbo showcased very similar performance between the EDP and SPP problems, while Claude Instant 1.2 exhibited similar performance for all these three problems. Yi-34b, Owen-14b, GPT 3.5 Turbo, and GPT 4 Turbo displayed remarkably high accuracy specifically for the SAS task. MPT-30b and Phi-1.5 showed very limited performance in identifying these three problems. 

Regarding NP-Complete complexity, there are several observations to highlight. Still, neither MPT-30b nor Phi-1.5 could deliver any identification for problems in the NP-Complete complexity. In the case of GCP-D and TSP-D problems, the performance of these models varied significantly. Phi-2, Vicuna-13b and GPT 4 Turbo outperformed in the GCP-D problem compared to TSP-D, whereas Claude Instant 1.2, Claude 2, and PaLM 2 exhibited better performance in TSP-D over GCP-D. On the other hand, models like Mistral-7b, Yi-34b, Qwen-14b, and GPT 3.5 Turbo showcased relatively similar performance between these two tasks. For the KSP task, only GPT 4 Turbo demonstrated promising performance, while the remaining models faltered. 

Considering NP-Hard complexity as the most intricate task set among the three (as evidenced in Figure \ref{Fig:complexities}), many of the examined models encountered challenges in identifying tasks within this complexity. For the GCP task, Mistral-7b, PaLM 2, GPT 3.5 Turbo, and GPT 4 Turbo exhibited some potential, while Vicuna-13b and Claude Instant 1.2 showed limited performance. For the TSP task, identification was observed only in Claude 2 and GPT 4 Turbo. Of all the investigated models, GPT 4 Turbo exhibited promise in identifying these three tasks within the NP-Hard complexity. However, the performance in GCP and TSP identification significantly surpassed that of the MSP task across these models. For the MSP task, only GPT 4 Turbo displayed some ability for identification, while with notably low accuracy. 

\subsection{Evaluating Benchmark Robustness}
\begin{figure}[h] 
    \centering
    \includegraphics[width= 1 \textwidth]{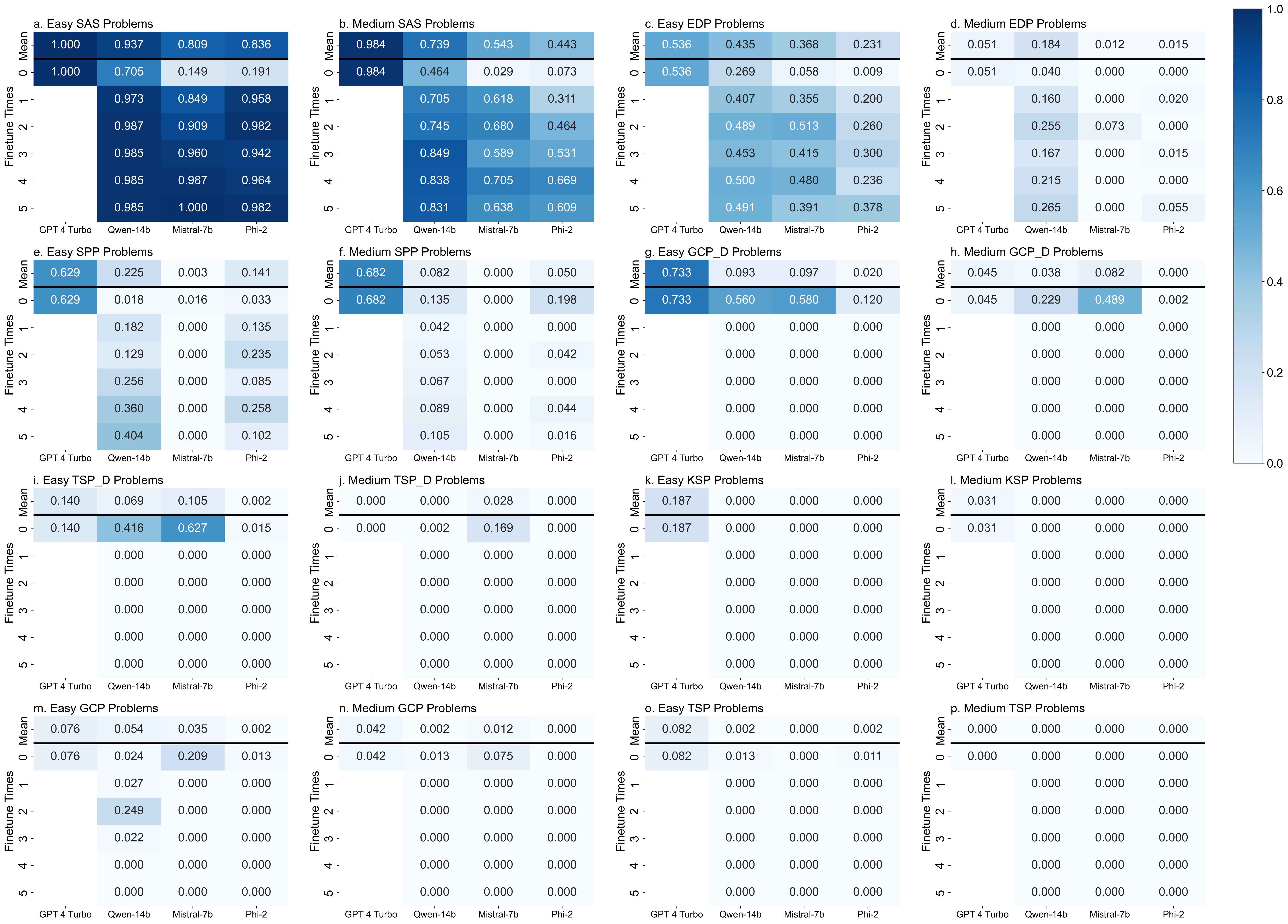}
    \caption{Model's robustness on different problems and difficulty levels.}
    \label{Fig:problems}
\end{figure}

In Experiment 2, we explore the robustness of benchmark against hacking attempts through a process of finetuning on pairs of question and gold answer. We experiment using 3 well-performing open-source models: Qwen-14b, Mistral-7b, and Phi-2 on two versions of benchmarks. Figure \ref{Fig:problems} presents the result\footnote{We do not present the result on MSP as this problem does not have a fixed solution and we do not conduct finetuning on it.}: each problem has two graph with one displaying evaluation results at difficulty levels 1-10 and one displaying evaluation results at difficulty levels 11-20. In each graph, the first row of indicate the accuracy mean of each model, averaged over outcomes at 5 finetuning checkpoints, ranging from tuning using zero (no finetuning) to five distinct benchmarks.

Our findings are twofold:
(1) While finetuning yields improvements in solving polynomial-time problems, its impact on the more complex NP-complete and NP-hard problems are negative. This suggests the inherent difficulty of hacking NP-complete, and potentially NP-hard, problems through the basic finetuning with question-and-answer approach. Manual annotation of the chain-of-thought, which is not provided in the benchmarks, could potentially enhance effectiveness, albeit with challenges in annotation.
(2) Finetuning appears beneficial for performance within the same difficulty level all P problems, yet shows limited out-of-distribution (OOD) adaptability and struggles to generalize to more difficult problems (as evidenced in graphs a and c) except SAS. For instance, Qwen-14b demonstrates notable proficiency on SPP challenges at levels 1-10 following finetuning; its performance is comparable to that of GPT-4. However, its performance significantly diminishes on SPP problems at levels 11-20, even underperforming compared to its unfinetuned checkpoint. This indicates that finetuning on these benchmarks can only benefit very simple questions such as SAS but could potentially impede generalization capabilities and renders finetuning hacking useless.

In conclusion, our benchmark is challenging to hack due to two primary factors: (1) the inherent complexity of NP-complete and NP-hard problems, which are difficult to learn solely from question-answer pairs, and (2) the propensity for P problems to become overfitted through finetuning on these pairs, while the real ``reasoning'' ability can be easily exposed by increasing the problem difficulty level. Based on these conclusions, we intend to periodically update our benchmarks strategetically with varying difficulty levels to minimize the potential for hacking.

\subsection{Effects of Few-shot Examples' Difficulty on Reasoning Ability Enhancement}
In Experiment 3, we focused on the tasks of SAS and EDP to investigate the nature of the in-context learning capabilities of LLMs. This experiment empirically distinguishes between ``learning'' and ``mimicking'' as exhibited by LLMs during in-context learning scenarios. Our findings also revealed a clear dichotomy in the approach to learning and generalization from examples between closed-source and open-source models.

For closed-source models, including GPT 4 Turbo, Claude 2, GPT 3.5 Turbo, PaLM 2, and Claude Instant 1.2, the results were notably close to the ideal scenario. We observed minimal variation in performance across different levels of difficulty in the examples provided. This consistency suggests that these models are not merely mimicking the solutions but are indeed learning the algorithmic skills presented in the context of the examples.

In contrast, the performance of open-source models, particularly Yi-34b and Mistral-7b, exhibits a clear pattern where the models generally generalize well from examples that are more challenging than the given question, yet they struggle to do so from simpler examples. Other open-source models display less distinct patterns, but a notable trend is still evident: these models demonstrate some capacity to generalize from more challenging to simpler questions, but they are less successful in generalizing from simpler to more complex questions. An exception is observed with the Phi-1.5 model in EDP, where it appears to generalize better from easier examples than from harder examples at certain difficulty levels. However, broadly speaking, none of the open-source models consistently learn from both harder and easier examples. The difficulty level significantly influences the models' performance, suggesting a tendency for these models to mimic patterns rather than engage in genuine learning from the context.

This phenomenon underscores that the differentiation between powerful closed-source and open-source models lies not only in their raw reasoning ability but also significantly in their capacity to learn from in-context examples. This insight highlights the importance of considering both reasoning and learning abilities when evaluating the effectiveness and potential applications of LLMs.

\begin{figure}[ht] 
    \centering
    \includegraphics[width = \textwidth]{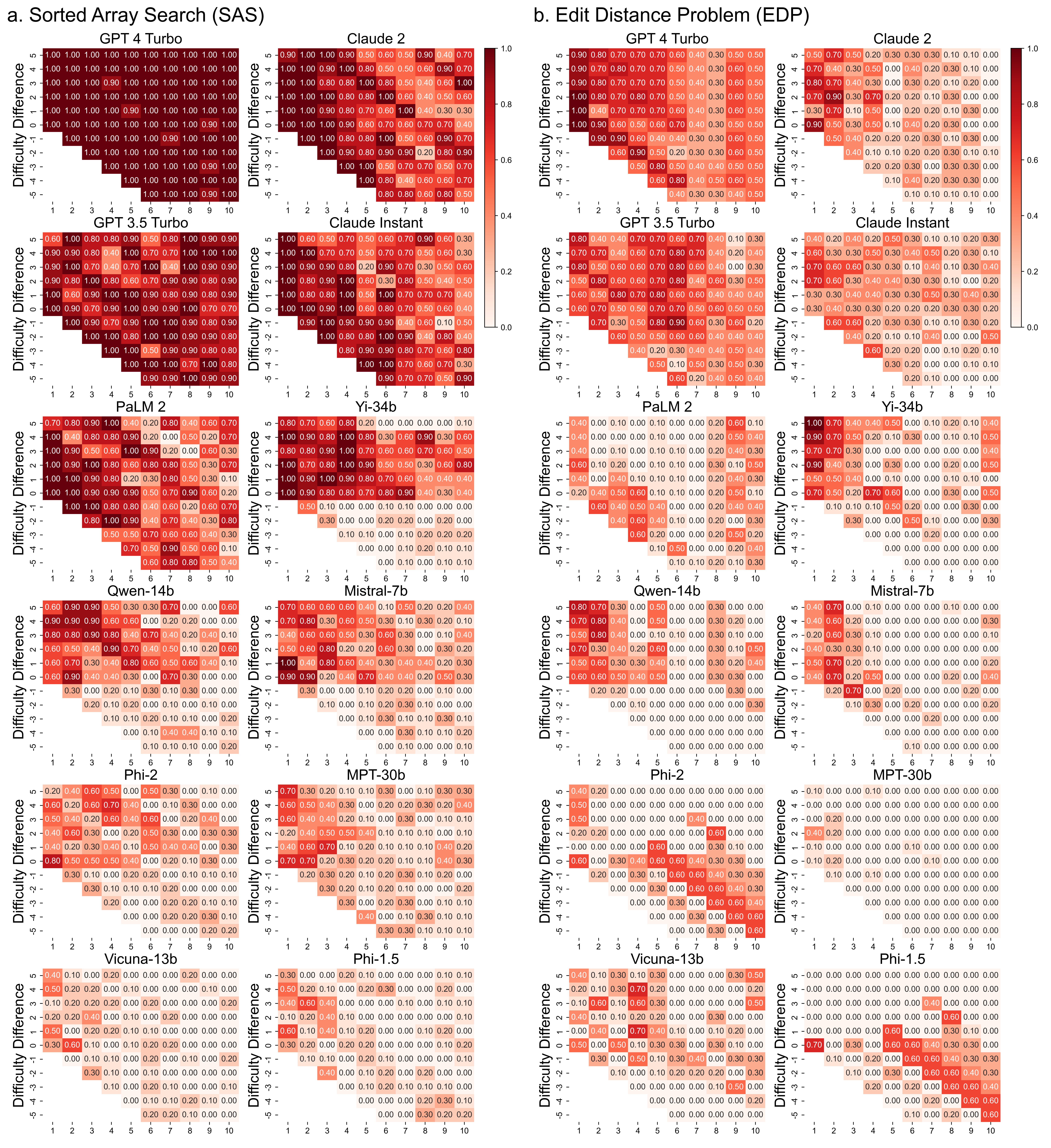}
    \caption{Heatmap of SAS and EDP task for each model.}
    \label{Fig:prompt_style_bsp}
\end{figure}

\begin{table}[ht]
\resizebox{\textwidth}{!}{
\begin{tabular}{lcccccccccccc} \toprule
\textbf{Accuracy} & \textbf{GPT 4 Turbo} & \textbf{Claude 2} & \textbf{GPT 3.5 Turbo} & \textbf{Claude Instant} & \textbf{PaLM 2} & \textbf{Yi-34b} & \textbf{Qwen-14b} & \textbf{Mistral-7b} & \textbf{Phi-2} & \textbf{MPT-30b} & \textbf{Vicuna-13b} & \textbf{Phi-1.5} \\ \midrule
Prompts on SAS \\ \midrule
Zeroshot & \textbf{1.000} & 0.445 & \textbf{0.942} & 0.442 & 0.416 & 0.620 & \textbf{0.706} & 0.149 & 0.191 & 0.000 & \textbf{0.113} & 0.000 \\ \midrule
Fewshot (-5) & 0.978          & 0.685          & 0.920         & 0.735          & 0.603          & 0.065    & 0.103      & 0.043    & 0.095         & 0.165          & 0.085  & \textbf{0.155} \\
Fewshot (-4) & \textbf{1.000} & 0.662          & 0.902         & 0.667          & 0.516          & 0.093    & 0.189      & 0.129  & 0.149        & 0.156          & 0.084  & 0.118          \\
Fewshot (-3) & 0.982          & 0.694          & 0.831         & \textbf{0.769} & 0.496          & 0.143         & 0.114      & 0.153    & 0.116      & 0.131          & 0.067  & 0.084          \\
Fewshot (-2) & \textbf{1.000} & \textbf{0.771} & 0.910         & 0.710          & 0.617          & 0.102         & 0.070    & 0.094    & 0.073      & 0.117          & 0.037  & 0.087          \\
Fewshot (-1) & 0.987          & 0.770          & 0.896         & 0.589          & 0.607          & 0.048      & 0.126    & 0.085    & 0.107      & 0.157          & 0.087  & 0.057          \\
Fewshot (0)  & 0.984          & 0.671          & 0.846         & 0.651          & \textbf{0.660} & 0.598    &0.255      & \textbf{0.413}       & 0.222       & 0.258          & 0.089  & 0.098          \\
Fewshot (1)  & 0.991          & 0.580          & 0.878         & 0.696          & 0.455          & 0.593    & 0.455     & 0.386   & \textbf{0.287}       & 0.233          & 0.055  & 0.109          \\
Fewshot (2)  & \textbf{1.000} & 0.675          & 0.829         & 0.587          & 0.656          & 0.647    & 0.444      & 0.296   & 0.260       & 0.175          & 0.067  & 0.056          \\
Fewshot (3)  & 0.993          & 0.736          & 0.800         & 0.598          & 0.489          & \textbf{0.662}     & 0.427  & 0.318    & 0.275      & 0.144          & 0.093  & 0.098          \\
Fewshot (4)  & \textbf{1.000} & 0.729          & 0.869         & 0.580          & 0.471          & 0.638    & 0.251      & 0.287     & 0.195     & \textbf{0.269} & 0.053  & 0.106          \\
Fewshot (5)  & \textbf{1.000} & 0.671          & 0.844         & 0.602          & 0.607          & 0.167    & 0.387      & 0.356     & 0.196     & 0.202          & 0.055  & 0.064  \\ \toprule
Prompts on EDP \\ \midrule
Zeroshot & 0.536 & 0.120 & 0.318 & 0.176 & 0.033 & 0.166 & \textbf{0.269} & 0.058 & 0.009 & 0.002 & 0.147 & 0.000 \\ \midrule
Fewshot (-5) & 0.387 & 0.075 & 0.417 & 0.048 & 0.170 & 0.000 & 0.000 & 0.015 & 0.210 & 0.000 & 0.000 & 0.205 \\ 
Fewshot (-4) & \textbf{0.556} & \textbf{0.209} & 0.367 & 0.102 & 0.207 & 0.000 & 0.000 & 0.000 & 0.300 & 0.000 & 0.044 & 0.284 \\ 
Fewshot (-3) & 0.500 & 0.178 & 0.386 & 0.167 & 0.235 & 0.029 & 0.000 &  0.029 & 0.327 & 0.000 & 0.108 & 0.331 \\ 
Fewshot (-2) & 0.462 & 0.173 & 0.479 & 0.210 & 0.208 & 0.146 & 0.065 & 0.090 & 0.329 & 0.000 & 0.154 & \textbf{0.335} \\ 
Fewshot (-1) & 0.485 & 0.200 & 0.513 & 0.246 & \textbf{0.289} & 0.135 & 0.069 & 0.098 & \textbf{0.348} & 0.011 & \textbf{0.248} & 0.328 \\ 
Fewshot (0) & 0.518 & \textbf{0.209} & \textbf{0.564} & 0.253 & 0.238 & \textbf{0.282} & 0.227 & \textbf{0.182} & 0.320 & \textbf{0.022} & 0.164 & 0.293 \\ 
Fewshot (1) & 0.535 & 0.184 & 0.535 & \textbf{0.355} & 0.205 & 0.089 & 0.266 & 0.089 & 0.115 & 0.013 & 0.160 & 0.115 \\ 
Fewshot (2) & 0.545 & \textbf{0.209} & 0.544 & 0.238 & 0.196 & 0.195 & 0.266 & 0.042 & 0.098 & 0.015 & 0.093 & 0.087 \\ 
Fewshot (3) & 0.536 & 0.189 & 0.449 & 0.315 & 0.182 & 0.127 & 0.140 & 0.067 & 0.060 & 0.007 & 0.191 & 0.051 \\ 
Fewshot (4) & 0.538 & 0.209 & 0.507 & 0.305 & 0.200 & 0.247 & 0.186 & 0.095 & 0.009 & 0.000 & 0.129 & 0.000 \\ 
Fewshot (5) & 0.531 & 0.205 & 0.449 & 0.244 & 0.167 & 0.271 & 0.146 & 0.055 & 0.015 & 0.009 & 0.202 & 0.000 \\ \bottomrule
\end{tabular}
}
\caption{Weighted accuracy of Zero-shot and Few-shot on SAS and EDP. The best performance for each column is highlighted with bold font (respectively for SAS and EDP).}
\end{table}


\begin{table}[ht]
\resizebox{\textwidth}{!}{
\begin{tabular}{lcccccccccccc} \toprule
\textbf{Failure Rate} & \textbf{GPT 4 Turbo} & \textbf{Claude 2} & \textbf{GPT 3.5 Turbo} & \textbf{Claude Instant} & \textbf{PaLM 2} & \textbf{Yi-34b} & \textbf{Qwen-14b} & \textbf{Mistral-7b} & \textbf{Phi-2} & \textbf{MPT-30b} & \textbf{Vicuna-13b} & \textbf{Phi-1.5} \\ \midrule
Prompts on SAS \\ \midrule
Zeroshot & 0.000 & 0.260 & 0.000 & 0.400 & 0.110 & 0.330 & 0.200 & 0.070 & 0.150 & 1.000 & 0.480 & 1.000 \\ \midrule
Fewshot (-5) & 0.000    & 0.060          & 0.020         & 0.000       & 0.000   & 0.940  & 0.900 & 0.480  & 0.920 & 0.160  & 0.640  & 0.860  \\
Fewshot (-4) & 0.000    & 0.033          & 0.000         & 0.000       & 0.017   & 0.917  & 0.817 & 0.617  & 0.867 & 0.117  & 0.633  & 0.900  \\
Fewshot (-3) & 0.000    & 0.057          & 0.057         & 0.014       & 0.000   & 0.871  & 0.886 & 0.471  & 0.886 & 0.043  & 0.600  & 0.914  \\
Fewshot (-2) & 0.000    & 0.075          & 0.025         & 0.000       & 0.013   & 0.888  & 0.913 & 0.538  & 0.900 & 0.063  & 0.613  & 0.888  \\
Fewshot (-1) & 0.000    & 0.044          & 0.022         & 0.000       & 0.011   & 0.911  & 0.856 & 0.622  & 0.878 & 0.078  & 0.589  & 0.933  \\
Fewshot (0)  & 0.000    & 0.060          & 0.020         & 0.000       & 0.020   & 0.300  & 0.640 & 0.380  & 0.670 & 0.040  & 0.540  & 0.880  \\
Fewshot (1)  & 0.000    & 0.040          & 0.020         & 0.010       & 0.010   & 0.290  & 0.500 & 0.420  & 0.700 & 0.060  & 0.520  & 0.830  \\
Fewshot (2)  & 0.000    & 0.040          & 0.040         & 0.000       & 0.010   & 0.290  & 0.510 & 0.440  & 0.710 & 0.030  & 0.640  & 0.910  \\
Fewshot (3)  & 0.000    & 0.020          & 0.050         & 0.010       & 0.030   & 0.280  & 0.450 & 0.460  & 0.670 & 0.030  & 0.650  & 0.830  \\
Fewshot (4)  & 0.000    & 0.050          & 0.030         & 0.000       & 0.040   & 0.280  & 0.570 & 0.530  & 0.700 & 0.080  & 0.630  & 0.850  \\
Fewshot (5)  & 0.000    & 0.050          & 0.040         & 0.000       & 0.020   & 0.680  & 0.520 &  0.440  & 0.740 & 0.050  & 0.650  & 0.900  \\ \toprule
Prompts on EDP \\ \midrule
Zeroshot & 0.000 & 0.000 & 0.000 & 0.000 & 0.440 & 0.000 & 0.000 & 0.040 & 0.000 & 0.960 & 0.160 & 0.950 \\ \midrule
Fewshot (-5) & 0.000 & 0.000 & 0.000 & 0.140 & 0.160 & 0.000 & 0.320 & 0.140 & 0.540 & 0.880 & 0.460 & 0.640 \\ 
Fewshot (-4) & 0.000 & 0.000 & 0.000 & 0.100 & 0.150 & 0.000 & 0.233 & 0.050 & 0.400 & 0.817 & 0.383 & 0.483 \\ 
Fewshot (-3) & 0.000 & 0.043 & 0.000 & 0.057 & 0.100 & 0.000 & 0.100 & 0.029 & 0.300 & 0.871 & 0.329 & 0.400 \\ 
Fewshot (-2) & 0.000 & 0.038 & 0.000 & 0.025 & 0.075 & 0.000 & 0.038 & 0.025 & 0.263 & 0.700 & 0.238 & 0.350 \\ 
Fewshot (-1) & 0.000 & 0.011 & 0.000 & 0.056 & 0.033 & 0.000 & 0.022 & 0.000 & 0.167 & 0.667 & 0.200 & 0.256 \\ 
Fewshot (0) & 0.000 & 0.020 & 0.000 & 0.060 & 0.000 & 0.000 & 0.040 & 0.000 & 0.130 & 0.730 & 0.190 & 0.200 \\ 
Fewshot (1) & 0.000 & 0.000 & 0.000 & 0.070 & 0.000 & 0.000 & 0.010 & 0.000 & 0.100 & 0.800 & 0.190 & 0.210 \\ 
Fewshot (2) & 0.000 & 0.040 & 0.000 & 0.050 & 0.000 & 0.000 & 0.020 & 0.000 & 0.000 & 0.750 & 0.090 & 0.100 \\ 
Fewshot (3) & 0.000 & 0.020 & 0.000 & 0.060 & 0.040 & 0.000 & 0.000 & 0.000 & 0.010 & 0.710 & 0.040 & 0.100 \\ 
Fewshot (4) & 0.000 & 0.030 & 0.000 & 0.030 & 0.000 & 0.000 & 0.000 & 0.000 & 0.000 &  0.810 & 0.000 & 0.000 \\ 
Fewshot (5) & 0.000 & 0.050 & 0.000 & 0.030 & 0.000 & 0.000 & 0.000 & 0.000 &  0.000 & 0.820 & 0.030 & 0.000 \\ \toprule
\end{tabular}
}
\caption{Weighted failure rate of Zero-shot and Few-shot on SAS and EDP.}
\end{table}

\section{Conclusions and Discussion}
In this study, we present a novel benchmark, \textbf{NPHardEval}, designed to rigorously evaluate LLMs' reasoning capabilities across a spectrum of complex tasks, up to the complexity class of NP-hard. By eschewing standard QA formats in favor of complex, logic-oriented problems, this benchmark aims to provide a more accurate measure of a model's reasoning prowess. This approach is crucial for developing LLMs capable of handling sophisticated, real-world tasks that demand high-level cognitive processing, steering the evaluation of LLMs from potentially ``useful" to fundamentally ``logical".

In addition to developing the benchmark, we compare different foundation models' reasoning ability across task complexity and experimented with different prompt styles to understand their in-context learnability. Our study reveals a notable disparity in performance between closed-source and open-source models not only on general reasoning ability but also the disparity between ``learning'' and ``mimicking''. 

With regard to models' performance across different complexity classes and difficulty levels, all models show decreased accuracy and increased failure rates as task complexity rises, with a marked performance decay at NP-Hard complexity levels. 
But the transition from P to NP-Complete complexity did not uniformly affect model performance; while some models showed little difference, others exhibited significant performance variations. Specifically, models like GPT 4 Turbo and Claude Instant 1.2 showed noteworthy performance shifts between these two complexity classes. Detailed performance analysis across specific tasks revealed that certain models had strengths in particular types of tasks within each complexity category, with a notable decline in model performance as they addressed more complex NP-Hard tasks. 

Finally, we used the tasks of SAS and EDP to understand how the difficulty of few-shot examples affects the in-context learning capabilities. Closed-source models like GPT 4 Turbo and Claude 2 demonstrated minimal performance variation and high consistency across different difficulty levels, suggesting a robust ability to learn algorithmic skills from examples. Conversely, open-source models showed varied adaptability, with some like Yi-34b and Mistral-7b performing well on more challenging examples but struggling with simpler ones. 

\subsection{Limitations}
While our study offers a novel approach to assessing the reasoning abilities of LLMs, it is paramount to reflect on the limitations of our current methodology to provide a comprehensive understanding and guide future research.

\paragraph{Task Complexity's Comparison} A significant limitation lies in the scope of our task selection and the definition of complexity within our benchmark. While we have delineated criteria for task selection in the appendix, a more resource-intensive approach could involve the inclusion of a larger variety of questions for each task type, enhancing the depth and breadth of our evaluation. Additionally, our current approach to defining complexity is based on a linear increment of weights. This simplistic weighting heuristic may not accurately represent the nuanced complexity increase in real-world tasks. More experimental work is needed to refine this approach and determine the most effective weight assignment that truly reflects the intricacies of task complexity.

\paragraph{Randomness} Another critical aspect to consider is the inherent randomness in the generation of responses by LLMs. This randomness can introduce variability in performance, making it challenging to draw consistent conclusions about a model's reasoning capabilities. \textbf{Notably, decision questions in the NP-complete level, including GCP-D and TSP-D, use true or false results as the evaluation criteria. Thus, it is hard to directly rule out the random positive cases, although the model may not go through a correct reasoning process, leading to potentially inflated performance.} Addressing this issue requires a more nuanced approach to evaluating responses, possibly through repeated trials or the incorporation of statistical methods to account for this variability.

\paragraph{Model Updates and Emergence} The fast-paced evolution of LLMs also presents a significant challenge. With the continuous version updates and emergence of advanced models like Gemini Ultra \cite{Gemini} and Phi-2 \cite{Phi-2}, as well as an increasing number of open-source options, the analysis based on our benchmark may quickly become outdated. Thus we will monitor and experiment on new models, together with the LLMs research community, to keep pace with these rapid developments is crucial for maintaining the relevance and applicability of our findings. This dynamic nature of the field necessitates a flexible and adaptable approach to benchmarking, where updates and revisions are integral to the evaluation process.

Future research should aim to expand the scope and depth of task selection, refine the complexity definition, account for generation randomness, and adapt to the evolving landscape of LLMs. Addressing these challenges will enhance the accuracy and relevance of our benchmark, contributing to the development of LLMs that are capable of sophisticated reasoning in complex, real-world scenarios.

\subsection{Research Outlook}
Our research outlook includes future investigations that can extend and enrich our understanding of the reasoning abilities of LLMs.

\paragraph{Fine-grained Time Complexity under Polynomial (P) with Big $\mathcal{O}$ notation} We will further the investigation of the P complexity class with fine-grained time complexity notation, the Big $\mathcal{O}$ notation. For example, the time complexity of SAS is $\mathcal{O}(\log n)$, while the time complexity of the Dijkstra algorithm, the solution to SPP, is $\mathcal{O}(V\log V + E)$ with Fibonacci heaps \cite{cormen2022introduction}. This approach will enable a detailed evaluation of models within the same complexity, proving a complement perspective to the current difficulty levels and enabling a possible cross-comparison among different tasks' difficulty levels. 

\paragraph{Self-correction for Reasoning} Another promising avenue is the enhancement of LLM reasoning abilities. A key strategy here is the implementation of iterative self-correction mechanisms. Pioneered by self-correction experiments in \cite{huang2023large, stechly2023gpt}, allowing LLMs to go through multiple rounds (e.g., ranging from 1 to 10) of self-correction, we can observe how the refinement process affects the accuracy and sophistication of their responses. This iterative process mimics human problem-solving, where multiple drafts and revisions lead to improved outcomes.

\paragraph{Multi-agent Systems for Reasoning} Moreover, exploring a multi-agent system \cite{wu2023autogen, chan2023chateval, ge2023openagi, ge2023llm} approach could significantly advance LLMs' reasoning abilities. In such a system, different LLM agents, each potentially specialized in certain types of reasoning or knowledge areas, collaborate to solve complex problems. This collaborative approach could mimic a team of experts, each contributing their expertise, leading to more comprehensive and nuanced solutions. It also opens the door to understanding how LLMs can interact and augment each other's capabilities, which is crucial for their application in real-world, multi-faceted problem-solving scenarios.

These future research directions hold the potential not only to deepen our understanding of the current capabilities and limitations of LLMs but also to drive forward the development of more sophisticated and reliable AI systems. By focusing on robustness testing and enhancing reasoning abilities through innovative methods like iterative self-correction and multi-agent systems, we can make significant strides towards realizing the full potential of LLMs in complex decision-making and problem-solving tasks.

\section{Acknowledgement}
We extend our sincere gratitude to Libby Hemphill for her invaluable support in this work. We also thank Jinkui Chi, who kindly contributed to the maintenance of the benchmark's code repository and user guidance. Additionally, we are grateful for the diverse feedback we received from Siqi Liu and many others, which illuminated our path towards enhancing the quality of this paper. 

\bibliographystyle{unsrt}  
\bibliography{references} 

\appendix
\section{Examples of Synthesized Data, the Corresponding Prompts, and LLMs' Outputs}
\label{Appendix:eg_experiment}

To further demonstrate the synthesized Data, the corresponding prompts, and LLMs' outputs, we choose two specific problems with different attributes, including the EDP problem from the P complexity class with linear data synthesis and the GCP problem from the NP-Hard complexity class with graph data synthesis. We provide the zero-shot prompt for these questions and the output based on the GPT 4 Turbo. The details of all prompts and results available at \url{https://github.com/casmlab/NPHardEval}.

\subsection{Example for P: EDP}
\begin{lstlisting}[language=HTML, caption=EDP Example]
# Synthesized Data
{
    "string_a": "cef",
    "string_b": "ccb"
}

# Prompt with Synthesized Data
The Edit Distance Problem (EDP) involves finding the minimum number of operations required to transform one string into another, where each operation is either an insertion, deletion, or substitution of a single character.
Find the minimum number of operations required to transform the first string cef into the second string ccb. The operations are insertion, deletion, and substitution of a single character, each requiring 1 edit operation.
Please provide the minimum number of operations required to transform the first string into the second string. Offer a brief explanation of your methodology. Keep your response concise and focused.
Enclose your output within <root></root> tags. Present your reasoning in <reasoning></reasoning> tags and the final minimum number of operations in <final_answer></final_answer> tags, like <final_answer>{'Operations': 'MINIMUM_NUMBER_OF_OPERATIONS'}</final_answer>.

# Output (formatted for easier reading)
<root>
    <reasoning>
        To transform \"cef\" into \"ccb\", we can follow these steps:\n
        1. Substitute 'e' with 'c' (cef -> ccf)\n
        2. Substitute 'f' with 'b' (ccf -> ccb)\n
        This results in two substitution operations.
    </reasoning>
    <final_answer>
        {"Operations": 2}
    </final_answer>
</root>
\end{lstlisting}

\subsection{Example for NP-Hard: GCP}
\begin{lstlisting}[language=HTML, caption=GCP Example]
# Synthesized Data
c This is a generated graph with chromatic number 2
p edge 6 4
e 1 6
e 2 6
e 3 4
e 3 5

# Prompt with Synthesized Data
Graph coloring refers to the problem of coloring vertices of a graph in such a way that no two adjacent vertices have the same color.
There are 6 vertices 1 to 6 in a graph. You may use 4 colors with alphabats from A, B, C,... to color the graph.
Please label every vertex, even if it is disconnected from the rest of the graph. Please provide each vertex's color. Do not skip any vertices. You should also provide very short step by step reasoning. Do not use multiple lines and try your best to save output tokens.
Your output should contain two parts enclosed by <root></root>. First, your step by step reasoning wraped by <reasoning></reasoning>. Second, the final output of all vertex numbers and their associated colors, wrapped by final_answer tag, like <final_answer>{0:'COLOR_1', 1:'COLOR_2', ...}</final_answer>.
The graph is below:
    Vertex 1 is connected to vertex 6.
    Vertex 2 is connected to vertex 6.
    Vertex 3 is connected to vertex 4.
    Vertex 3 is connected to vertex 5.

# Output (formatted for easier reading)
<root>
    <reasoning>
        Start with vertex 1, color it A; color adjacent vertex 3 with B; vertex 2 can be A as it's not adjacent to 1; vertex 4 connected to 2 must be different, use B; vertex 5 connected to both 2 and 3, use C; vertex 6 connected to 3 and 4, use A.
    </reasoning>
    <final_answer>
        {1:'A', 2:'A', 3:'B', 4:'B', 5:'C', 6:'A'}
    </final_answer>
</root>
\end{lstlisting}

\section{Choices of Problems}
\label{Appnedix:task_exclude}

In the benchmark, we exclude calculation-only (math intensive) tasks for each of the complexity classes, due to the overlap with already exist benchmarks and the known uncertainty of LLMs' math ability. For other reasoning, we provide detailed explanations and highlight them in bold.

\subsection{Excluded P problems}
\paragraph{Prime Number Determination} Using algorithms like AKS primality test to determine if a given number is prime. Reason: Math-intensive.
\paragraph{Solving Linear Equations} Finding solutions for a system of linear equations. Reason: Math-intensive.
\paragraph{Maximum Flow Problem} Finding the maximum flow from a source node to a sink node in a flow network. A flow network is a directed graph $G = (V, E)$ where each edge $(u, v) \in E$ has a capacity $c(u, v)$ and flow $f(u, v)$, with a designated source $s$ and sink $t$. The objective is to maximize the total flow from $s$ to $t$ under the constraints that the flow on an edge does not exceed its capacity and the incoming flow is equal to the outgoing flow for every vertex except $s$ and $t$. Reason: \textbf{Most open source algorithms cannot follow the question and the prompt to provide outputs with mostly correct formats}.

\subsection{Excluded NP-Complete problems}
\paragraph{3-SAT Problem} Deciding whether a given Boolean formula in conjunctive normal form with three literals per clause is satisfiable. Reason: Math-intensive.

\subsection{Excluded NP-hard problems}
\paragraph{Integer Linear Programming} Finding the best integer solution for a set of linear equations and inequalities. Reason: Math-intensive.

\end{document}